\documentclass{sig-alternate-per}
\usepackage[utf8]{inputenc}
\usepackage{microtype}
\usepackage{graphicx}
\usepackage{booktabs} 
\usepackage[font={footnotesize}]{caption}
\usepackage[font={footnotesize}]{subcaption}
\usepackage{algorithm}
\usepackage{algorithmic}
\usepackage{hyperref}

\usepackage{color}

\usepackage{nicefrac}       
\usepackage{microtype}      
\usepackage{float}
\usepackage{csquotes}
\usepackage[outdir=./]{epstopdf}

\usepackage{wrapfig}
\usepackage{tikz}
\usetikzlibrary {arrows.meta}
\usepackage{thmtools, thm-restate}
\usepackage{bbm}

\usepackage{graphicx}
\usepackage{booktabs}
\usepackage{caption}
\usepackage{comment}
\usepackage{amsmath}
\usepackage{amssymb}
\usepackage{amsfonts}
\usepackage{mathletters} 
\usepackage{thmtools, thm-restate}

\usepackage{pgfplots}
\pgfplotsset{compat=1.15}
\usepackage{mathrsfs}
\usetikzlibrary{arrows}
\title{Better than the Best: Gradient-based Improper Reinforcement Learning for Network Scheduling}
\author{Mohammadi Zaki, Avi Mohan, Aditya Gopalan and Shie Mannor}
\begin{document}

\maketitle

\begin{abstract}
We consider the problem of scheduling in constrained queueing networks with a view to minimizing packet delay. 
%
We formulate a novel top down approach to scheduling where, given an unknown network and a set of scheduling policies, we use a policy gradient based reinforcement learning algorithm that produces a scheduler that performs better than the available atomic policies. We derive convergence results and analyze finite time performance of the algorithm. Simulation results show that the algorithm performs well even when the arrival rates are nonstationary and can stabilize the system even when the constituent policies are unstable. Link to paper: \url{https://arxiv.org/pdf/2102.08201.pdf}
\end{abstract}
\section{Introduction}
The design of communication networks has traditionally involved fine-grained modeling of traffic and network characteristics, followed by devising scheduling and routing protocols optimized for the models. This constituted a \enquote{bottom-up} approach, with resource allocation algorithms being tightly coupled to network model assumptions, and worked very well with the relatively simple requirements of yesteryear networks comprising mostly homogeneous traffic sources. It has yielded a readily available storehouse of design principles and rules of thumb that can be used to generate with little effort, a menu of schedulers with desirable properties, e.g., MaxWeight and variants \cite{tassiulas-ephremides92stability-queueing, shakkottai2002scheduling}. 

Modern communication systems, on the other hand, are becoming increasingly complex, and are required to handle multiple types of traffic with widely varying characteristics (such as arrival rates and service times). This, coupled with the need for rapid network deployment, render such a granular model-based and bottom-up approach infeasible. 
At the same time, we would prefer to retain the favorable principles underlying the design of existing scheduling algorithms and not give them up altogether. %
In this regard, this paper advocates a novel \enquote{top-down} approach to designing effective and adaptable scheduling strategies, in which, given an unknown network and a set of scheduling policies, we aim to learn a scheduler that is either better than all the atomic policies or as good as the best atomic policy (which is a priori unknown for the network setting at hand). We achieve this using a gradient-based optimization algorithm over an improper mixture class of policies constructed using the given base controllers. 
We employ tools from recent analyses of policy gradient methods to derive convergence results and analyze finite-time performance of the proposed algorithm in this new setting. Simulation results show that the algorithm performs well even when the arrival rates are nonstationary, and can stabilize the system even when the constituent policies are unstable.

Related work is discussed in detail in Sec.~1.1 of our tech report \cite{zaki-etal21improper-learning-gradient-policy} and is omitted here due to paucity of space.

\section{System Setting}
To describe our approach in detail, we focus on the well-known setting of a single server attending to $N$ queues in discrete time, where for example, each queue models packets waiting on a communication link. We note, however, that our algorithmic framework applies more generally to policy optimization for any Markov decision process (MDP), including one with continuous state/action spaces, as long as a set of policies is given for it (please refer to \cite{zaki-etal21improper-learning-gradient-policy} for the complete formulation). 
The server decides which queues are to be scheduled for service in each slot, based on service constraints (e.g., at most 1 queue to be scheduled each time). The indicator random variable $D_i(t)$ denotes whether queue $i$ is scheduled in slot $t$ or not. For concreteness, we also assume (1) IID Bernoulli($\lambda_i$) arrivals $\{A_i(t)\}_{t \in \mathbb{N}}$ to each queue $i$, (2) deterministic, single-packet service for each queue when scheduled, and (3) a scheduling constraint of at most $1$ queue per time slot.
Note, however, that our policy optimization approach extends to general arrival processes or interference graphs (i.e., scheduling constraints). Hence, the queue lengths evolve as $
    Q_i(t+1) = \left(Q_i(t)-D_i(t)\right)^+ + A_i(t+1),~i\in[N], ~t \in \mathbb{N}$,
where  $(x)^+:=\max\{0,x\},~\forall~x\in\mathbb{R}$. %
Note that the arrival rates $\boldsymbol{\lambda}=[\lambda_1,\cdots,\lambda_N]$ are a priori {\color{blue}unknown} to the scheduler/learner. 
The learner (i.e., scheduling algorithm at the server) needs to decide which of the $N$ queues it intends to serve in a given slot. The server's decision at each slot can be denoted by the vector $\mathbf{D}(t) = (D_i(t))_{i\in N}$  taking values in the action space $\mathcal{A}:=\left\lbrace[0,\cdots,0],[1,0,\cdots,0],[0,0,\cdots,1]\right\rbrace,$ where a \enquote{$1$} denotes service and a \enquote{$0$} denotes lack thereof.
Let ${H}_t$ denote the state-action history (historical trajectory) until time $t,$ and $\mathcal{P}(\mathcal{A})$ the space of all probability distributions on $\mathcal{A}.$ We aim to find a policy $\pi=[\pi_1,\pi_2,\cdots]$, where $\pi_t: {H}_t\rightarrow\mathcal{P}\left(\mathcal{A}\right)$, to minimize the $\infty$-horizon  discounted system backlog given by 
\begin{equation}
    J_\pi(\mathbf{q}):=\mathbb{E}^\pi\left[ \sum_{t=0}^{\infty}\gamma^t\sum_{i=1}^N Q_i(t) \given \mathbf{Q}(0) = \mathbf{q} \right].
    \label{eqn:queueing-discounted-cost}
\end{equation}
Note that we are using system backlogs (queue lengths) as a proxy for packet delays as is commonly done; a more finer performance criterion involving the actual packet delays can also be optimized if the MDP is suitably redefined. Any policy $\pi$ with $J_\pi(\mathbf{Q}(0))<\infty,~\forall\mathbf{Q}(0)\in\mathbb{Z}_+^2$ is said to be \emph{\color{blue}stabilizing} (or, equivalently, a \emph{stable} policy). The capacity region \cite{tassiulas-ephremides92stability-queueing} of this network can be seen to be $\left\lbrace\boldsymbol{\lambda}\in\Real^N_+\mid\sum_{i\in[N]}\lambda_i<1\right\rbrace$. 

This problem can be viewed as one of finding an $\infty$-horizon, $\gamma$-discounted \emph{\color{blue}reward} optimal policy in the MDP with state space $\mathcal{S}=\mathbb{N}^N,$ i.e., all possible values of the queue lengths $\mathbf{Q}(t) \equiv (Q_i(t))_{i \in N}$,  action space $\mathcal{P}(\mathcal{A})$, single stage reward $r(\mathbf{Q}(t),\mathbf{D}(t))=-\sum_{i=1}^NQ_i(t)$, and an appropriately defined probability transition kernel $\tP$ following the Bernoulli arrival process  \cite{zaki-etal21improper-learning-gradient-policy}. In keeping with standard reinforcement learning parlance, we will refer to the negative discounted system backlog $-J_\pi(\cdot)$ as the  \emph{value function} $V^{\pi}\left(\cdot \right)$ of policy $\pi$, to be maximized over policies $\pi$. Moreover, due to the Markov nature of the system, we consider only policies that depend on the current state, i.e., $\pi_t: \mathbf{Q}(t) \to \mathcal{P}(\mathcal{A})$.

{\bfseries{Improper Learning.}} We assume that we are provided with a finite number of controllers/policies $\mathcal{C}:=\{K_1,\cdots,K_M\}$. We aim to identify the best policy for the given queueing network within a class, i.e.,
\begin{equation}\label{eq:main optimization problem}
\pi^* = \argmin\limits_{\pi\in \mathcal{I}_{soft}(\mathcal{C})} V^{\pi}(\rho),     
\end{equation}
where $\mathcal{I}_{soft}(\mathcal{C})$ is a parameterized, improper, policy class that we define as follows.


{\bfseries{The Softmax Policy Class.}} Each policy in the softmax policy class $\mathcal{I}_{soft}(\mathcal{C})$ is parameterized by weights $\theta:=[\theta_1,\cdots,\theta_M] \in \mathbb{R}^M$. 
The policy $\pi_{\theta}\in\mathcal{I}_{soft}(\mathcal{C})$, given a state $s \in \mathcal{S}$, plays an action by (a) first choosing a controller drawn from {\color{blue}$\softmax(\theta)$}, i.e., the probability of choosing controller~$K_m$ is given by,
\begin{equation}
  \pi_\theta(m):= \frac{e^{\theta_m}}{\sum\limits_{m'=1}^M e^{\theta_{m'}}},  
  \label{eqn:softmax-defn}
\end{equation}
and (b) then choosing an action by applying the sampled controller at the state $s$.
Note, therefore, that in every round, our algorithm decides which action to apply only \emph{through} the controller sampled in the first step of that round. In the rest of the paper, we will deal exclusively with a fixed base policy class $\mathcal{C}$ and the resultant  $\mathcal{I}_{soft}(\mathcal{C})$. We use the notation $\pi_{\theta}(a|s)$ for any $a\in \cA$ and $s\in \cS$ to denote the probability with which the softmax policy $\pi_\theta$, as defined above, chooses action $a$ in state $s$. Hence, we have that for any $\theta \in \mathbb{R}^m$,
\begin{equation}\label{eq:policy action selection criteria}
    \pi_{\theta}(a|s) = \sum\limits_{m=1}^{M}\pi_{\theta}(m)K_m(s,a), 
\end{equation}
where $K_m(s,a)$ is the probability with which the policy $K_m$ plays $a$ in state $s$. Since we deal with gradient-based methods in the sequel, we define the \emph{\color{blue}value gradient} of policy $\pi_{\theta}\in\mathcal{I}_{soft},$ by $\nabla_{\theta}V^{\pi_\theta}$.

\section{Scheduling via Policy Gradients}
\textbf{The Policy Gradient Approach.} The Policy Gradient (PG) method has, following stunning success with applications such as game playing, has become a cornerstone of reinforcement learning \cite{sutton-barto18reinforcementBOOK}. In general, PG methods involve parameterizing the control policy and optimizing the parameter using a gradient ascent algorithm of the form
\begin{equation}
    \boldsymbol{\theta}_{t+1} = \boldsymbol{\theta}_t + \eta \nabla_{\theta_t}V^{\pi_{\theta_t}}
\end{equation}
When the value function and its gradient are computable in closed form, we propose an algorithm, \emph{\color{blue}SoftMax PG}, that provably converges to the best parameter, $\boldsymbol{\theta}^*$, within $\mathcal{I}_{soft}(\mathcal{C}).$
\begin{algorithm}[tb]
   \caption{Softmax Policy Gradient (SoftMax PG)}
   \label{alg:mainPolicyGradMDP}
\begin{algorithmic}
   \STATE {\bfseries Input:} learning rate $\eta>0$, initial state distribution $\mu$
   \STATE Initialize each $\theta^1_m=1$, for all $m\in [M]$, $s_1\sim \mu$
   \FOR{$t=1$ {\bfseries to} $T$}
   \STATE Choose controller $m_t\sim \pi_t$.
   \STATE Play action $a_t \sim K_{m_t}(s_{t},:)$.
   \STATE Observe $s_{t+1}\sim \tP(.|s_t,a_t)$.
   \STATE Update: $\theta_{t+1} = \theta_{t} + \eta. \nabla_{\theta_t}V^{\pi_{\theta_t}}$.
   \ENDFOR
\end{algorithmic}
\end{algorithm}
\subsection{Convergence of SoftMax PG}
\begin{figure*}[ht]
\centering
    \begin{subfigure}[t]{.23\textwidth}
        \includegraphics[scale = 0.29]{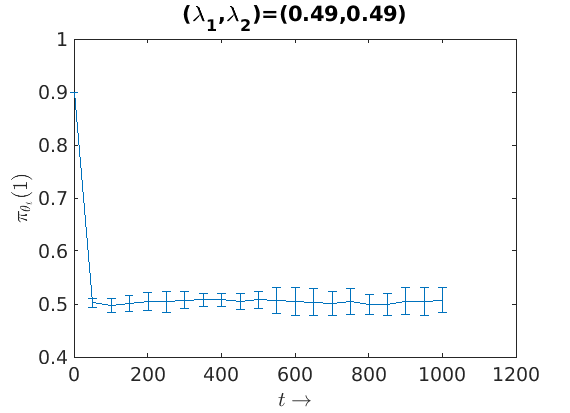}
        \caption{$\boldsymbol{\lambda} = (0.49,0.49)$,  $\mathcal{C} =\{\text{`always serve 1', `always serve 2'}\}$ }
        \label{subfig:equal arrival rate}
    \end{subfigure}%
    \hfill
    \begin{subfigure}[t]{.23\textwidth}
        \includegraphics[scale = 0.29]{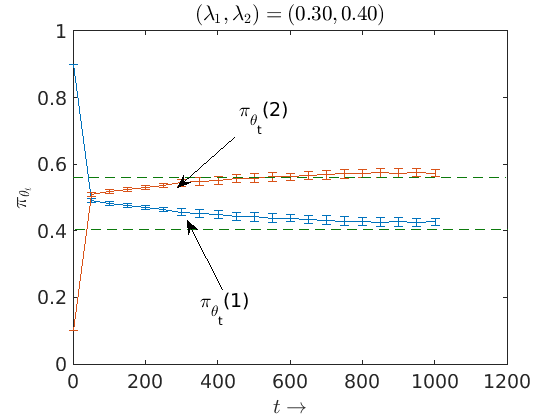}
        \caption{$\boldsymbol{\lambda} = (0.3,0.4)$,   $ \mathcal{C} = \{\text{`always serve 1', `always serve 2'}\}$ }
        \label{subfig:unequal arrival rate}
    \end{subfigure}%
    \hfill
        \begin{subfigure}[t]{.23\textwidth}
        \includegraphics[scale = 0.29]{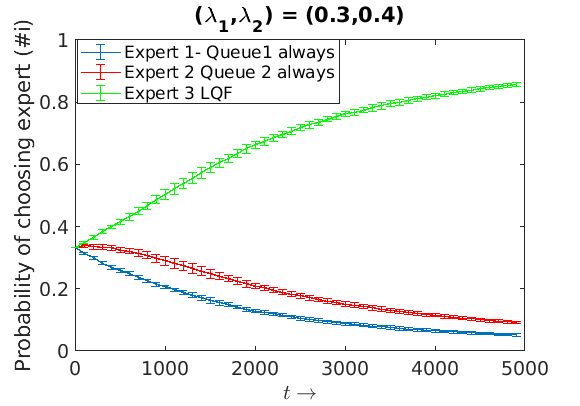}
        \caption{$\boldsymbol{\lambda} = (0.3,0.4)$,   $ \mathcal{C} = \{\text{`serve 1', `serve 2', LQF}\}$}
        \label{subfig:3experts}
    \end{subfigure}%
    \hfill
    \begin{subfigure}[t]{.23\textwidth}
        \includegraphics[scale = 0.29]{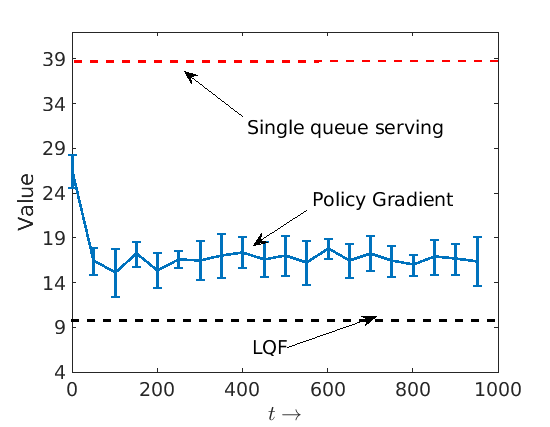}
        \caption{Expected discounted queue length cost for \ref{subfig:3experts}
        }
        \label{subfig:Value function comparisom}
    \end{subfigure}%
\caption{\small Softmax policy gradient with gradient estimation finds the best mixture policy for various base policies in a $2$-queue network.}
\label{fig:simulations}
\end{figure*}
For a given policy $\pi$ and initial state distribution $\mu,$ the quantity $d_\mu^\pi(\cdot):=\mathbb{E}_{\mathbf{Q}_0\sim \mu}\left[(1-\gamma)\sum\limits_{t=0}^\infty \prob{\mathbf{Q}_t=\cdot\given \mathbf{Q}_o,\pi,\tP} \right]$, defines a distribution over $\mathcal{S}$, called the \emph{discounted state visitation measure.}
\begin{restatable}[Rate of Convergence]{theorem}{maintheorem}\label{thm:convergence of policy gradient}
Let $\abs{\mathcal{S}\times\mathcal{A}}<\infty$, $\mathbf{Q}(0)\sim\mu$ and assume that the scheduler is provided with $M$ stationary controllers.
With $\{\theta_t \}_{t\geq 1}$ generated as in Algorithm \ref{alg:mainPolicyGradMDP} and using a learning rate $\eta = \frac{\left(1-\gamma\right)^2}{7\gamma^2+4\gamma+5}$, for all $t\geq 1$,
\[V^{\pi_{\theta_t}}(\rho) - V^{\pi^*}(\rho) \leq {\color{blue}\frac{1}{t}} M \left(\frac{7\gamma^2+4\gamma+5}{c^2(1-\gamma)^3}\right)\norm{\frac{d_\mu^{\pi^*}}{\mu}}_{\infty}^2 \norm{\frac{1}{\mu}}_\infty.  \]
Here, $c = \inf_{t \geq 1} \min \{\pi_{\theta_t}(m): m \in [M], \pi^*(m) > 0 \}$ is the minimum probability that Algorithm \ref{alg:mainPolicyGradMDP} puts on the controllers on which the best mixture $\pi^*$ is supported.
\end{restatable}
%

\subsection{Estimating Value Gradients}
In most RL problems, neither value functions nor value gradients are available in closed-form. To expanding the applicability of SoftMax PG to such situations, we propose a gradient estimation subroutine called \emph{\color{blue} GradEst} (Algorithm \ref{alg:gradEst}). This uses a combination of (1) rollouts to estimate the value of the current (improper) policy and (2) a stochastic perturbation-based approach to estimate its value gradient. 

Specifically, in order to estimate the value gradient, we use the approach of Flaxman et al \cite{Flaxman05}, noting that for $V:\Real^M\to\Real$, the gradient $
    \nabla V(\theta) \approx \expect{\left(V(\theta+\alpha. u)-V(\theta) \right)u}.\frac{M}{\alpha}$. 
where $\alpha \in (0,1)$.
If $u$ is chosen to be uniformly random on unit sphere, the second term is zero, i.e., \\$ \expect{\left(V(\theta+\alpha u)-V(\theta) \right)u}.\frac{M}{\alpha} = \expect{\left(V(\theta+\alpha u) \right)u}.\frac{M}{\alpha}.$

The expression above requires evaluation of the value function at the point $(\theta+\alpha.u)$. Since the value function may not be explicitly computable, we employ rollouts for its evaluation. 

\begin{algorithm}[tb]
   \caption{GradEst}
   \label{alg:gradEst}
\begin{algorithmic}
   \STATE {\bfseries Input:} Policy parameters $\theta$, parameter $\alpha>0$.
   \FOR{$i=1$ {\bfseries to} {\tt \#runs}}
   \STATE $u^i\sim Unif(\mathbb{S}^{M-1}).$
   \STATE $\theta_\alpha = \theta+\alpha.u^i$
   \STATE $\pi_\alpha = \softmax(\theta_\alpha)$
   \FOR{$l=1$ {\bfseries to} {\tt \#rollouts}}
   \STATE Generate trajectory according to the policy $\pi_\alpha:$ $(s_0,a_0,r_0,s_1,a_1,r_1,\ldots, s_{lt},a_{lt},r_{lt})$
   \STATE {\tt reward}$(l)=\sum\limits_{j=0}^{lt}\gamma^jr_j$
   \ENDFOR
   \STATE  {\tt mr}$(i) = {\tt mean}(\text{{\tt reward}})$
   \ENDFOR
   \STATE {\tt GradValue} $= \frac{1}{\text{{\tt \#runs}}}\sum\limits_{i=1}^{\text{{\tt \#runs}}}\text{{\tt mr}}(i).u^i.\frac{M}{\alpha}.$
   \STATE {\bfseries return} {\tt GradValue}
\end{algorithmic}
\end{algorithm}

\section{Experiments}
In this section, we show the efficacy of our policy gradient algorithm through simulations in multiple scenarios. 
We study the performance of softmax PG with gradient estimation (GradEst) over two different settings: (1) when the packet arrival rates, and therefore the optimal controller, are fixed and (2) where they are time varying. In all our experiments, we consider a system with $N=2$ queues and a fully connected interference graph. We provide all details about hyperparameters in  \cite[Sec.~E]{zaki-etal21improper-learning-gradient-policy}. 
\subsection{Constant Arrival Rates}
In this case, we first simulate a network where the the optimal policy ($\pi_{\boldsymbol{\theta^*}}$) is a strict improper combination of the available controllers and later, a network where it is at a corner point, i.e., one of the available controllers itself is optimal. Our simulations show that in both the cases, softmax PG converges to the correct controller distribution in $\mathcal{I}_{soft}$.  

The scheduler is given two base/atomic controllers $\mathcal{C}:=\{K_1,K_2\}$, i.e. $M=2$. Controller $K_i$ serves Queue~$i$ with probability $1$, $i=1,2$. As can be seen in Fig.~\ref{subfig:equal arrival rate} when $\boldsymbol{\lambda}=[0.49,0.49]$, softmax PG converges to the improper mixture policy that serves each queue independently with probability $[0.5,0.5]$, which is the delay-optimal controller in $\mathcal{I}_{soft}(\mathcal{C})$. Interestingly, the mixture stabilizes the system whereas both base controllers lead to instability because of insufficient service to some queue. Fig.~\ref{subfig:unequal arrival rate} shows that with unequal arrival rates too, Softmax-PG with GradEst quickly converges to the correct improper combination. 

Fig.~\ref{subfig:Value function comparisom} shows the evolution of the value function (system backlog) of GradEst (blue) compared with those of the base controllers (red) and the \emph{Longest Queue First} policy (LQF) which, as the name suggests, always serves the longest queue in the system (black). LQF, like any work-conserving policy, is known to be delay optimal \cite{LQF2016}. 

Finally, Fig.~\ref{subfig:3experts} shows the result of the second experimental setting with three atomic controllers, one of which is delay optimal. The first two are $K_1,K_2$ as before and the third controller, $K_3$, is LQF. Notice that $K_1,K_2$ are both queue length-agnostic, meaning they could attempt to serve empty queues as well. LQF, on the other hand, always and only serves nonempty queues. Hence, in this case the optimal policy is attained at one of the corner points, i.e., $[0,0,1]$. The plot shows the GradEst converging to the correct point on the simplex.

\subsection{Time-varying Arrival Rates}
\begin{figure}[tb]
        \centering
        \includegraphics[scale=0.22]{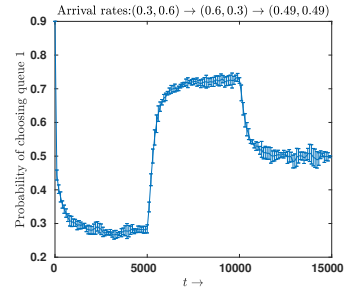}
        \caption{\small Plot showing that GradEst \emph{adapts} to varying arrival rates over time.}
        \label{fig:nonstationaryexample}
        \vspace{-0.50cm}
    \end{figure}

We consider a modification to the system in Sec.~6 wherein the arrival rates $\boldsymbol{\lambda}$ to the two queues vary over time (adversarially). In particular, $\boldsymbol{\lambda}$ varies from $(0.3,0.6)\to (0.6,0.3)\to(0.49,0.49)$. Our PG algorithm successfully \emph{tracks} this change and \emph{adapts} to the optimal improper stationary policies in each case as shown in  Fig.~\ref{fig:nonstationaryexample}. In all three cases a mixed controller is optimal, and is successfully tracked by our PG algorithm.
\section{Conclusion}
Our results show that our new improper learning algorithmic framework is able to efficiently learn optimal mixtures of given policies. This paves the way for (a) building more refined theory towards understanding the convergence behavior of such schemes, and (b) benchmarking it more extensively in diverse RL settings including problems of robotic control, computer game playing, etc. This will form the subject of future work.  

\bibliographystyle{acm}
\bibliography{main}
\end{document}